\documentclass[letterpaper, 10 pt, conference]{ieeeconf}  

\IEEEoverridecommandlockouts                              

\usepackage{graphics} 
    \DeclareGraphicsExtensions{.jpg,.pdf,.mps,.png} 
    \graphicspath{{fig/} {./}} 
\usepackage{epsfig} 
\usepackage{amsmath} 
\usepackage{amssymb}  
\usepackage{bm} 
\usepackage{flushend} 

\newcommand{\fig}[1]{Fig.~\ref{#1}}

\newcommand{\tab}[1]{Table~\ref{#1}}

\def\epsgaiji#1{\leavevmode\kern-0.025zw\raise-.37zh\hbox{%
  \epsfile{file=#1,width=1.05zw}}\kern-0.025zw}
\newcommand{\MARU}[1]{{\ooalign{\hfil#1\/\hfil\crcr\raise.167ex\hbox{\mathhexbox20D}}}}

\usepackage{array}

\usepackage{siunitx} 
\usepackage{gensymb} 
\usepackage{booktabs} 
\usepackage{multirow} 
\usepackage{tabularx}    
\usepackage{stfloats}    

\let\labelindent\relax 
\usepackage{enumitem} 

\usepackage{pgfplots}
\pgfplotsset{compat=newest}
\usetikzlibrary{plotmarks}
\usetikzlibrary{arrows.meta}
\usepgfplotslibrary{patchplots}
\usepackage{xcolor}
\usepackage[hyphens]{url}
\usepackage[hidelinks]{hyperref} 

\usepackage{grffile}
\pgfplotsset{plot coordinates/math parser=false}
\newlength\fwidth
\newlength\fheight

\usepackage{cite}

\title{\LARGE \bf
A Pin-Array Structured Climbing Robot for Stable Locomotion \\on Steep Rocky Terrain
}
\author{Keita Nagaoka$^{1\dag}$, Kentaro Uno$^{1\dag}$, and Kazuya Yoshida$^{1}$
\thanks{$^{*}$This work is supported by JSPS KAKENHI Grant Number JP23K13281.}
\thanks{$^{1}$The authors are with the Space Robotics Lab. (SRL) in Department of Aerospace Engineering, Graduate School of Engineering, Tohoku University,
        Sendai 980-8579, Japan.
        }%
\thanks{$^{\dag}$\textit{These authors contributed equally to this work. Corresponding author is Kentaro Uno}{ (Email: {\tt unoken@tohoku.ac.jp}).}}
}
\begin{document}

\maketitle
\thispagestyle{empty}
\pagestyle{empty}


\begin{abstract}
Climbing robots face significant challenges when navigating unstructured environments, where reliable attachment to irregular surfaces is critical. We present a novel mobile climbing robot equipped with compliant pin-array structured grippers that passively conform to surface irregularities, ensuring stable ground gripping without the need for complicated sensing or control. Each pin features a vertically split design, combining an elastic element with a metal spine to enable mechanical interlocking with microscale surface features. Statistical modeling and experimental validation indicate that variability in individual pin forces and contact numbers are the primary sources of grasping uncertainty. The robot demonstrated robust and stable locomotion in indoor tests on inclined walls (10$^{\circ}$--30$^{\circ}$) and in outdoor tests on natural rocky terrain. This work highlights that a design emphasizing passive compliance and mechanical redundancy provides a practical and robust solution for real-world climbing robots while minimizing control complexity.
\end{abstract}


\section{Introduction} 

Grippers capable of reliably handling diverse shapes have applications in industrial automation, exploration, and disaster-response robotics. A prominent approach is the state-of-the-art gripper using the jamming transition, which encloses granular particles in an elastic membrane and varies the internal pressure to adjust stiffness, enabling conformity and secure grasping~\cite{nature_jamming}. Many promising improved versions for more advanced usage scenarios~\cite{6142115,Fujita03062018} have also been studied. Furthermore, stiffness-variable grippers using magnetorheological fluids have also been proposed~\cite{PETTERSSON2010332,8287269, mgf_gripper}.
\begin{figure}[t]
  \centering
  \includegraphics[width=\linewidth]{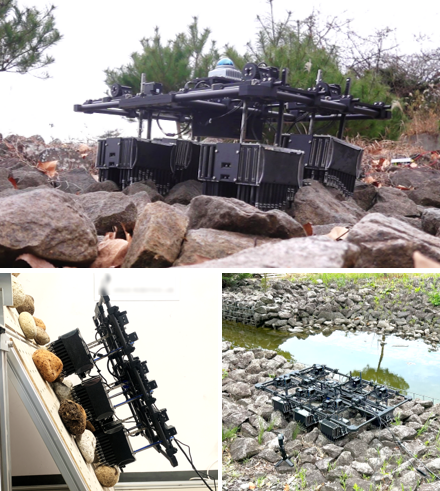}
  \caption{The pin-array unit passively adapts to irregular terrain shapes, enabling stable ground-gripping locomotion (top). The robot demonstrated stable adhesion performance on steeply inclined rocky terrain (bottom left) and was also deployed in a natural rough landscape (bottom right).}
  \label{fig1}
\end{figure}
Another approach is the pin-array gripper, inspired by contour gauges, in which pins passively deform to match object contours. A representative example is the Omnigripper developed by Scott {\it et al.}~\cite{omnigripper}, which employs two passively driven pin arrays that conform to the target and then lock in place to generate friction for stable grasping. Variations include pins that rotate individually to generate friction~\cite{pin_rotate, mo_pin_array}, radial arrangements around the gripper center~\cite{crpa_gripper}, and inflatable balloons mounted at the pin tips~\cite{BalloonPinArrayGripper}. Kato {\it et al.} further developed a claw-tipped model with elastic elements at the pin ends, enabling grasping of rocky surfaces by generating pressing forces and embedding the claws into the terrain~\cite{kato2022pin-array}.

Most prior studies, however, have focused on gripper-level grasping rather than on integration into robotic systems. Pinbot~\cite{pinbot}, proposed by Noh {\it et al.}, is an example of a mobile robot utilizing three pin-array units for locomotion; however, it lacks true grasping capability and has not been validated for high-load climbing performance. On the other hand, existing legged climbing robots~\cite{MARCBot2024, scaler, uno2021hubrobo, limbero} typically employ ground-gripping end-effectors that are limited to engaging convex surfaces, thereby restricting their adaptability to continuously irregular natural terrains.

To address this challenge, we integrated such pin-array modules into a novel mobile robot with gripping and climbing capabilities on rocky and steep terrain. To the authors' best knowledge, this paper serves as the initial report to study the climbing capability of a robot that exploits pin-array structures for such mobility. The contributions of this paper are highlighted as follows:
\begin{itemize}
    \item Designed and realized a novel climbing robot that integrates a pin-array gripper module, enabling stable locomotion on rocky and steep terrain.
    \item Developed a predictive model of the pin-array structured gripping performance and validated it through systematic experimental testing, providing insights for future scaling and design optimization.
    \item Demonstrated robust and repeatable locomotion capability in both controlled indoor experiments and challenging outdoor natural terrain, confirming the practical applicability of the proposed mechanism.
\end{itemize}

\section{Pin-Array Structured Gripper}
A pin-array gripper, a core technology of the proposed climbing robot, was developed based on the work by Kato {\it et al.}~\cite{kato2022pin-array}. In this study, the gripper was further improved by increasing the pin density, enabling enhanced gripping performance per unit, suitable for mobile robotic applications.
\begin{figure*}[t]
    \centering
    \includegraphics[width=\linewidth]{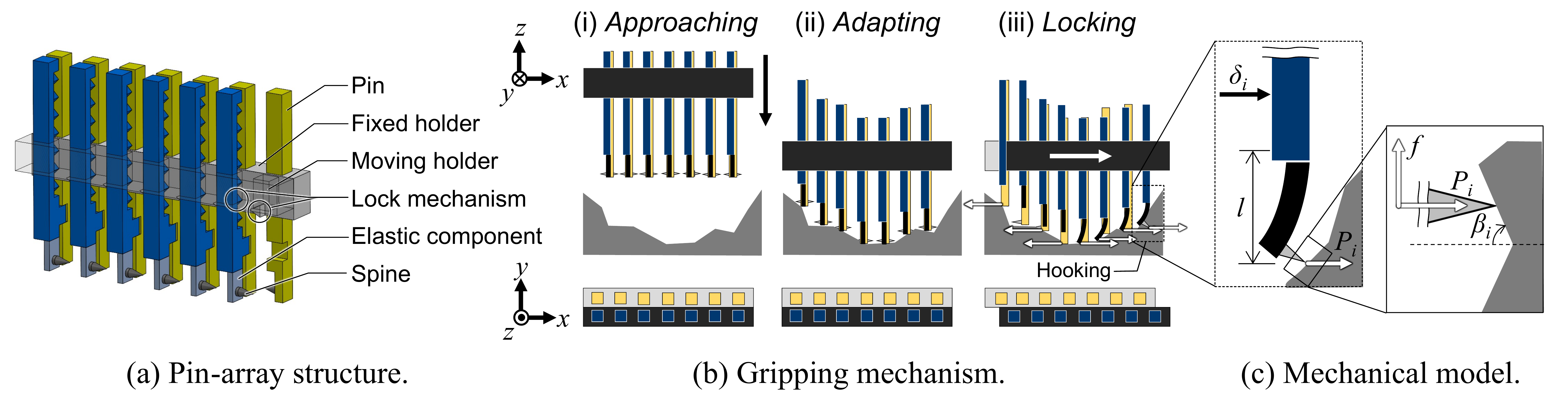}
    \caption{Overview of the pin-array structured gripping technology: (a) hardware structure, (b) gripping mechanism, and (c) mechanical model of contact. A concave shape is exemplified as a terrain surface in (b); however, the same mechanism also works for convex shapes for gripping.}
    \label{pin-array_gripper_mechanism}
\end{figure*}
\subsection{Principle}

The mechanism shown in \fig{pin-array_gripper_mechanism}(a) consists of a pin-array capable of passive vertical motion and its holders. Each pin is vertically split into two parts; the lower section (tip) is equipped with an elastic element acting as a leaf spring and a metal spike. The split pins are housed in holders that constrain their horizontal motion. The rear holder (the \textit{fixed holder} in \fig{pin-array_gripper_mechanism}) is rigidly attached to the mechanism body, while the front holder (the \textit{moving holder} in \fig{pin-array_gripper_mechanism}) is horizontally actuated. Each pin is also guided by concave or convex guides to ensure its split parts move synchronously.

The sequence of operations for grasping uneven surfaces is illustrated in \fig{pin-array_gripper_mechanism}(b). The grasping procedure is as follows:
\begin{enumerate}[label=(\roman*)]
    \item Approaching: Approach the surface approximately perpendicular to the terrain.
    \item Adapting: Press the mechanism against the surface, allowing the pin array to conform to the terrain.
    \item Locking: Drive the movable holder horizontally using a motor.
\end{enumerate}
Through this motion, the elastic elements generate a restoring force that presses the spines against the surface, producing a holding force via mechanical interlocking. Therefore, the pin-array structured gripper generates engagement forces that pinch from the outside toward the inside when engaging convex surfaces, and forces that push outward from the inside when engaging concave surfaces. This shape-independent gripping is achieved by the identical operation of the gripper using a single actuator, without any prior sensing.

The holding force of the mechanism is determined by the engagement of the spines at the tips of the pins with the surface, as well as the pressing force exerted by the spines on the surface. This pressing force arises from the restoring force of the elastic component at the base of each pin, which is induced by the horizontal motion of the moving holder.
As illustrated in Fig.~\ref{pin-array_gripper_mechanism}(c), each elastic element can be modeled as a cantilever beam. The pressing force $P_i$ is determined by the horizontal displacement $\delta_i$ of the moving holder after the spine makes contact with the surface, according to the formula:
\begin{eqnarray}
    P_i = \frac{3\delta_iEI}{l^3}
    \label{eq:2_1}
\end{eqnarray}
where $E$ is the Young's modulus of the elastic element, $I$ is its second moment of area, and $l$ is the effective length from the fixed end of the elastic element to the tip of the spine.

\subsection{Gripper Design and Development}
\begin{figure}[t]
  \centering
  \includegraphics[width=.9\linewidth]{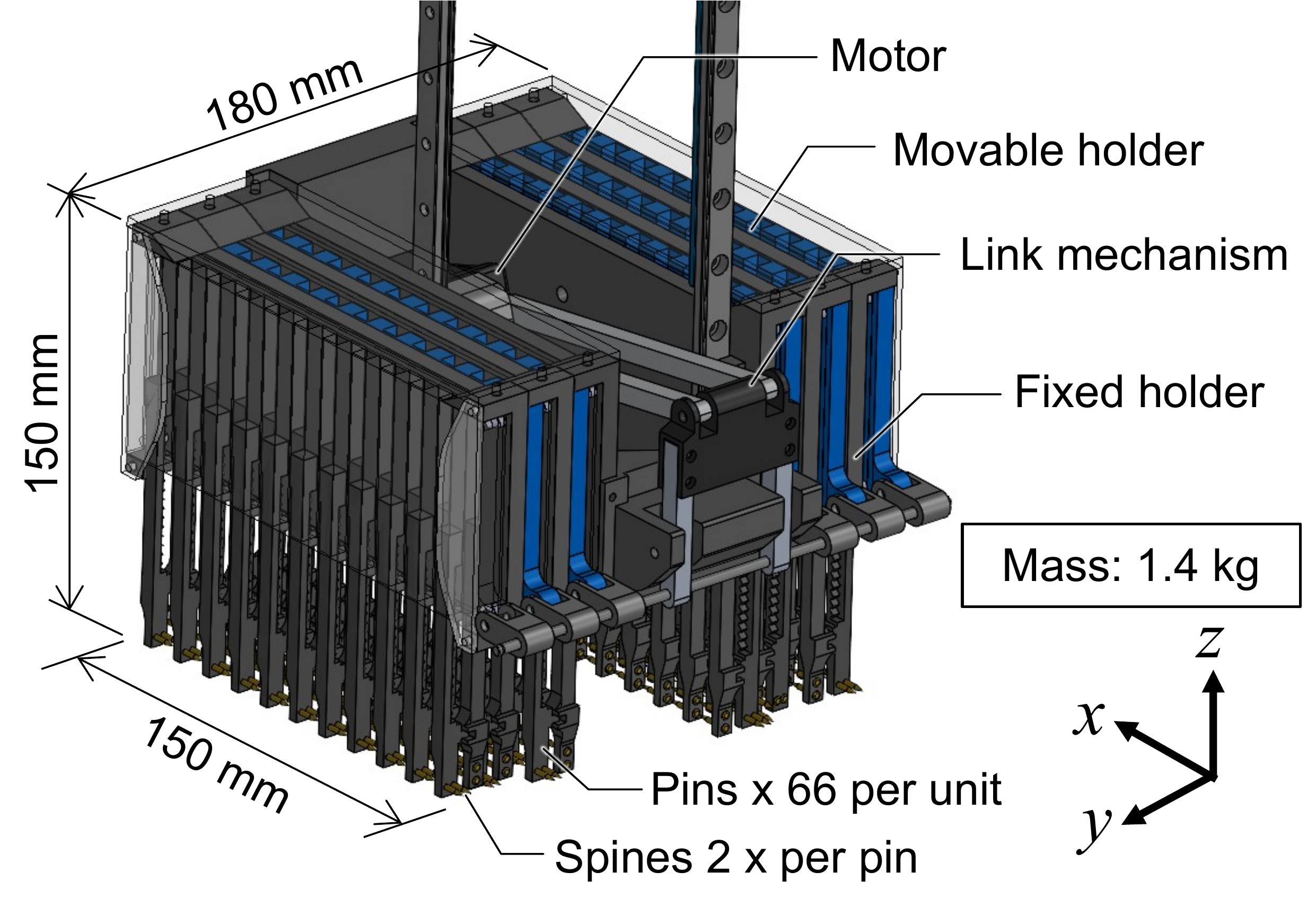}
  \caption{Pin-array gripper unit design. Sixty-six pins are installed and driven by a single actuator.}
  \label{fig_developed_gripper}
\end{figure}
The developed pin-array gripper unit is shown in \fig{fig_developed_gripper}. 
The mechanism has dimensions of 150\;mm in height (with pins fully extended), 180\;mm in width, and 150\;mm in depth, with a total mass of 1.40\;kg. 
The width and depth were specifically set to 150\;mm to match the 10--20\;cm scale of the rocks in the indoor testbed, ensuring that the unit can sufficiently cover the surface area of a single rock or bridge multiple contact points.
Each holder contains 11 pins, and the basic configuration of two holders stacked in three layers results in a total of 66 pins.

The pins and holders are 3D-printed from carbon fiber–reinforced polyphthalamide (PPA-CF) and polylactic acid (PLA), respectively. The elastic elements (leaf springs) are made of polycarbonate, while the pin tips are fitted with brass nails to provide mechanical engagement. A locking mechanism between the fixed holder and the pins constrains their vertical ($z$-axis) motion during grasping. This prevents the gripped object from slipping out along with the pins.

To ensure gravity-independent surface conformity and minimize contact forces, low-stiffness compression springs (spring coefficient: $k = 0.098$;N/m) are installed on the upper side of each pin within the fixed holder.

The gripper's holding force improves with higher pin density and a broader horizontal displacement range. However, expanding the displacement typically requires larger, high-stiffness components, which limits density. To address this, the pin array was optimized within a 150 mm actuation range---scaled to match the 10--20 cm rocky terrain---to maximize contact probability. To accommodate the 5 mm minimum diameter of commercially available compression springs, the pin diameter was reduced to its physical limit. This refinement achieved a 1.5-fold increase in pin density per unit compared to previous work (Fig.~\ref{fig_developed_gripper}). By doubling the number of basic units from three to six, a total of $N=132$ pins was attained, maximizing spatial resolution under hardware constraints. To support this density, PPA-CF was selected for its 5.2 times higher flexural modulus than ABS, providing the necessary rigidity while maintaining a lightweight structure. Additionally, the actuator was integrated within the housing to minimize size and ensure a full operational range.


\subsection{Holding Force Evaluation}
\begin{figure}[t]
  \centering
  \includegraphics[width=\linewidth]{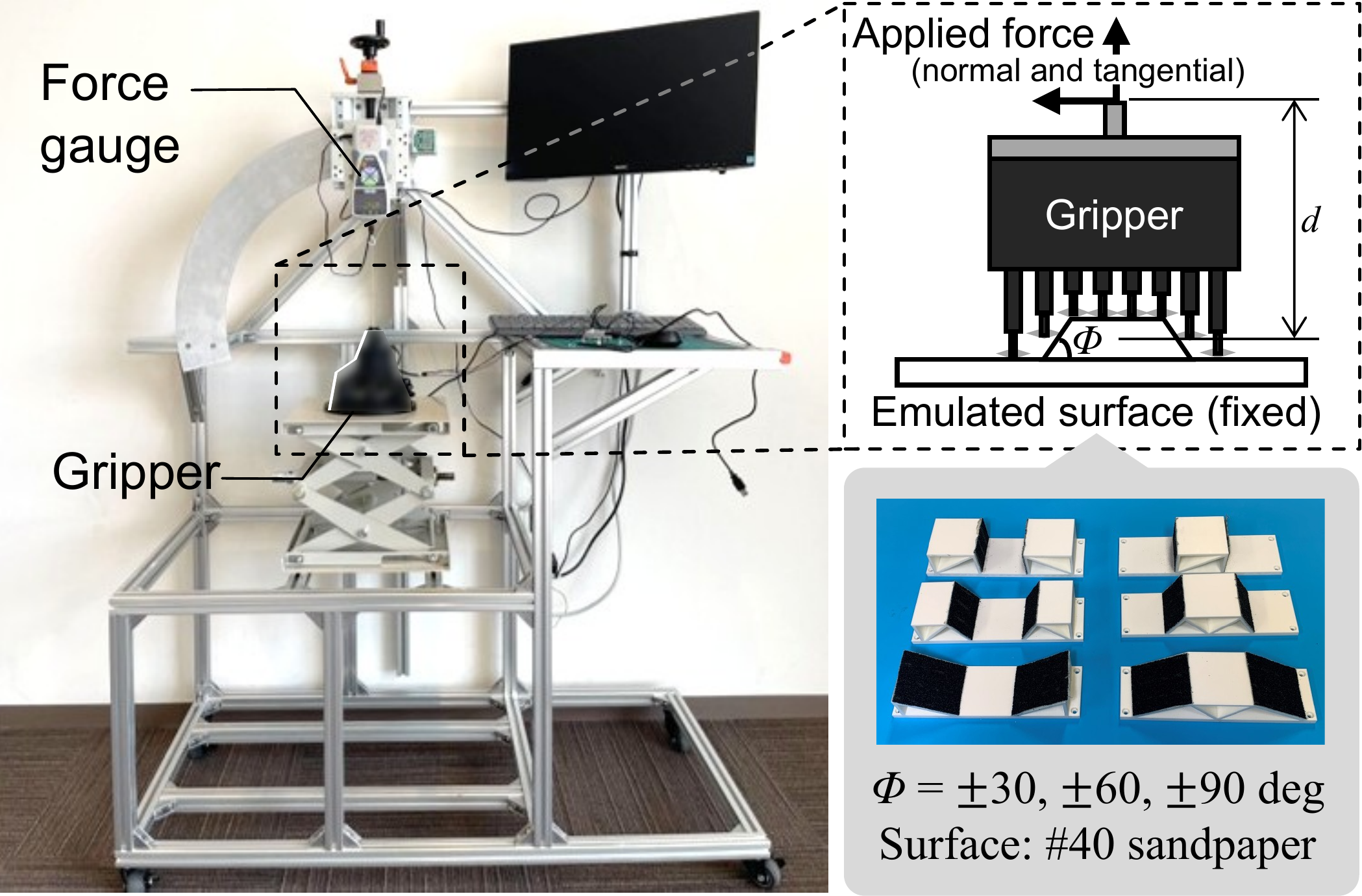}
  \caption{Holding force measurement setup. The pin-array gripper's engaging performance was tested on various convex and concave objects with different ${\it \Phi} = \pm30^{\circ}, \pm60^{\circ}, \pm90^{\circ}$, while the height of the force application point from the terrain was kept constant at $d = 0.14$\;m.}
  \label{fig:holding_force_measurement_testing_setup}
\end{figure}

The pin-array gripper developed in this study was evaluated in terms of its holding force against various types of emulated surface models.
The holding force testing setup
and emulated surfaces are shown in \fig{fig:holding_force_measurement_testing_setup}. The device allows the gripper to be pulled at arbitrary angles from vertical to horizontal, with forces recorded by an integrated gauge. Six types of emulated surfaces with different slope angles ${\it \Phi}$ of $\pm 30^{\circ}$, $\pm 60^{\circ}$, and $\pm 90^{\circ}$---where positive ${\it \Phi}$ indicates convex shapes and negative ${\it \Phi}$ indicates concave shapes---were prepared, with the sloped regions covered by \#40 sandpaper to emulate the surfaces. Two pulling directions were tested: normal and tangential. For the normal direction, the holding force was calculated at the instant of complete detachment minus the gripper’s self-weight; for the tangential direction, it was taken directly from the gauge. Each condition was repeated ten times for statistical evaluation.


\tab{tab:holding_force_measument_result} summarizes the results of the gripping force experiments. In the normal direction, the lowest holding force, 7.39\;N, was obtained at $\phi = 30^{\circ}$, while the overall average was 34.1\;N. In the tangential direction, the minimum value, 6.11\;N, likewise occurred at $\phi = 30^{\circ}$, with an overall average of 21.1\;N.

\begin{table}[b]
  \centering
  \caption{Summary of holding force test results. Minimum and average holding forces were obtained in both the normal and tangential directions at various surface angles {\it $\phi$}.}
  \label{tab:holding_force_measument_result}
    \begin{tabular}{lcc}
        \hline
        Direction & Minimum Value [N] & Overall Average [N] \\
        \hline
        Normal     & 7.39 & 34.1 \\
        Tangential & 6.11 & 21.1 \\
        \hline
    \end{tabular}
\end{table}
\begin{figure}[t]
  \centering
  \includegraphics[width=\linewidth]{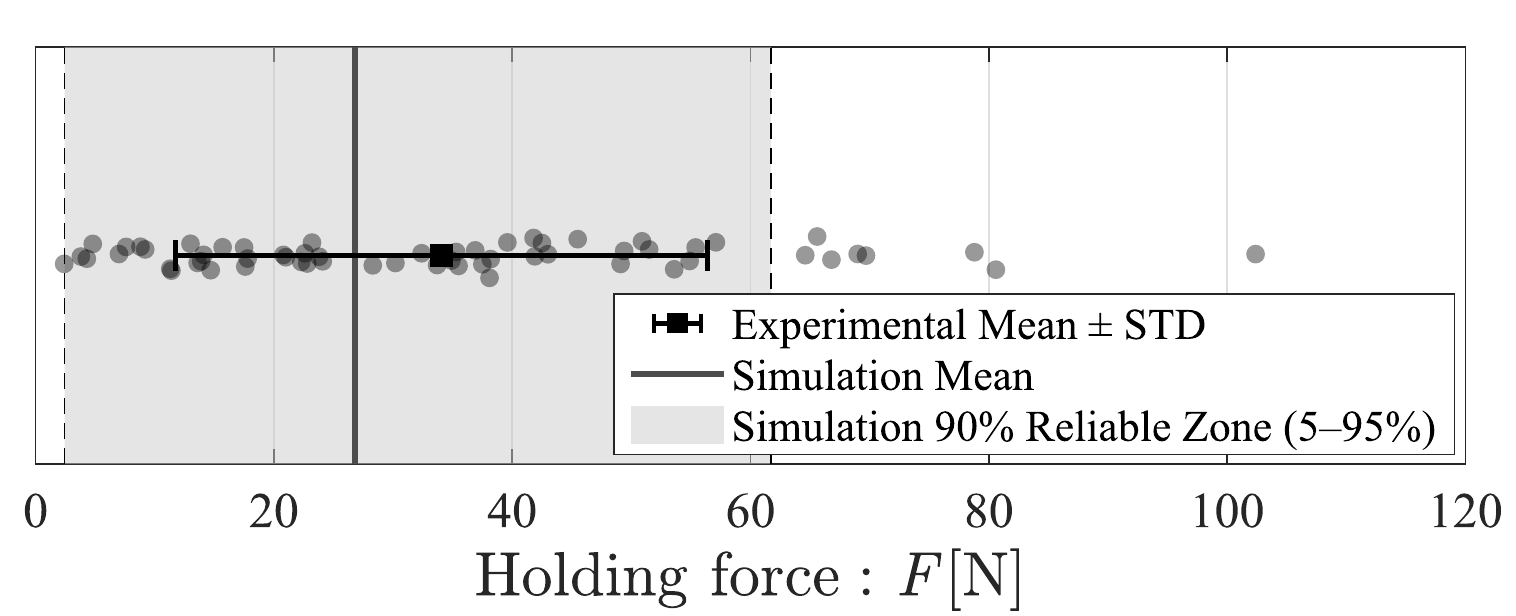}
  \caption{Simulation results validating the mechanical model of the pin-array gripper. The points correspond to the measurements from the holding force evaluation tests and are largely encompassed within the confidence interval.}
  \label{fig:grasping_force_simulation}
\end{figure}

\subsection{Holding Force Modeling}
Modeling the achievable holding force of the pin-array structured gripper is essential for extending this design to different scales. Due to the substantial mechanical variability of individual pins, exact analytical modeling of the gripping force is impractical. Accordingly, in this study, each variable in Eq.~\eqref{eq:2_1} is treated probabilistically, and a statistical approach is adopted to capture the overall behavior of the pin-array gripper.

\subsubsection{Modeling}
In the proposed mechanism, the holding force of a single pin can be expressed as
\begin{equation}
    F_i = \mu'_i P_i
\end{equation}
where \(P_i\) denotes the normal force applied to pin \(i\), and \(\mu'_i\) is the apparent static friction coefficient. Unlike a conventional Coulomb friction coefficient, \(\mu'_i\) accounts for the physical engagement of the pin tip with micro-asperities on the surface~\cite{asbeck_microspine}. It depends on the inclination angle \(\beta_i\) of the surface asperities and the true friction coefficient \(\mu\), and is given by
\begin{equation}
    \mu'_i = \frac{\mu + \tan\beta_i}{1 - \mu\tan\beta_i}
\end{equation}
This concept theoretically explains why the holding force of the gripper cannot be captured by a simple Coulomb friction model. Assuming that the total load \(F\) is evenly distributed among the \(n\) pins in contact with the surface, the total holding force can be expressed as
\begin{equation}
    F = \sum_{i=1}^{n} F_i = \sum_{i=1}^{n} \mu'_i P_i.
\end{equation}

However, the gripping performance of each individual pin exhibits considerable mechanical variability, making exact modeling of the holding force impractical. Therefore, a statistical approach is employed by introducing variability into each parameter of the above equation. In particular, probabilistic assumptions are imposed on the number of contacting pins, the normal forces of the pins, and the apparent friction coefficients. The total holding force is then evaluated using Monte Carlo simulation, allowing the behavior of the pin-array gripper to be statistically modeled and compared with experimental measurements.
\begin{itemize}
    \item Number of contacting pins ($n$):
        Out of the total $N_{\text{pins}}=132$ pins, each pin is assumed to independently contact the surface with a constant probability of $p_{\text{contact}}=0.1$. Under this assumption, the number of contacting pins $n$ follows a binomial distribution $B(N_{\text{pins}}, p_{\text{contact}})$.
    \item Pushing force ($P_i$):
        The pushing force acting on each contacting pin, $P_i$, is assumed to follow a gamma distribution $\Gamma(k, \theta)$ to represent variability in the applied force. The distribution parameters $k$ and $\theta$ are determined from the mean $\mu$ and variance $\sigma^2$ using the method of moments:
\begin{align}
    k &= \frac{\mu^2}{\sigma^2}, &
    \theta &= \frac{\sigma^2}{\mu}
\end{align}
        The mean and variance are determined based on prior experimental data and preliminary tests, under the physical assumption that the normal force lies within the range 0--5\;N.
    \item Apparent coefficient of friction ($\mu'i$):
        Assumed to follow a uniform distribution $\mu'i$ is $U(0.4, 3.0)$, to account for variations in surface and pin tip conditions.
\end{itemize}

\subsubsection{Validation of model}
Fig.~\ref{fig:grasping_force_simulation} compares the measured experimental grasping forces with the results from our probabilistic Monte Carlo simulations. The simulations predicted a 90\% confidence interval of $[19.8~\text{N}, 62.9~\text{N}]$ with a mean of 39.8\;N. The simulated confidence interval (shaded area) encompasses the variability of the experimental data across most inclination angles.

These results suggest that the primary sources of uncertainty in the gripper's grasping force are the variability in individual pin forces and the probabilistic number of contacts—the core assumptions of our model. However, some mean experimental values exceed the simulation's upper bound. This is likely because larger-than-expected local angles between the pin tips and surface asperities caused the apparent friction coefficient, \(\mu'_i\), to exceed its assumed maximum of 3.0.

The contact probability $p_{contact} = 0.1$ was identified as a representative parameter that captures the stochastic nature of the terrain engagement. Given the high packing density of the pins and the potential for mechanical interference between adjacent units, an effective engagement rate of 10\% is physically plausible for the porous and irregular surfaces evaluated. Crucially, the fact that the simulated variance consistently aligns with the experimental data distribution across the entire range of inclination angles validates this parameter identification. This demonstrates that $p_{contact} = 0.1$ effectively encapsulates the underlying physical interaction between the pin-array and the rough terrain, providing a reliable basis for predicting grasping performance.

In summary, our simulation model captures the essential characteristics of the complex force generation mechanism and supports the validity of the experimental data. The results confirm that a statistical approach is an effective means to model the grasping force of a multiple pin-array gripper. The developed mechanical model is useful for predicting the achievable gripping performance for future designs of different scales, as well as the ground-gripping capability of a climbing robot using this mechanism.

\section{Robot development}
\begin{figure}[t]
  \centering
  \includegraphics[width=\linewidth]{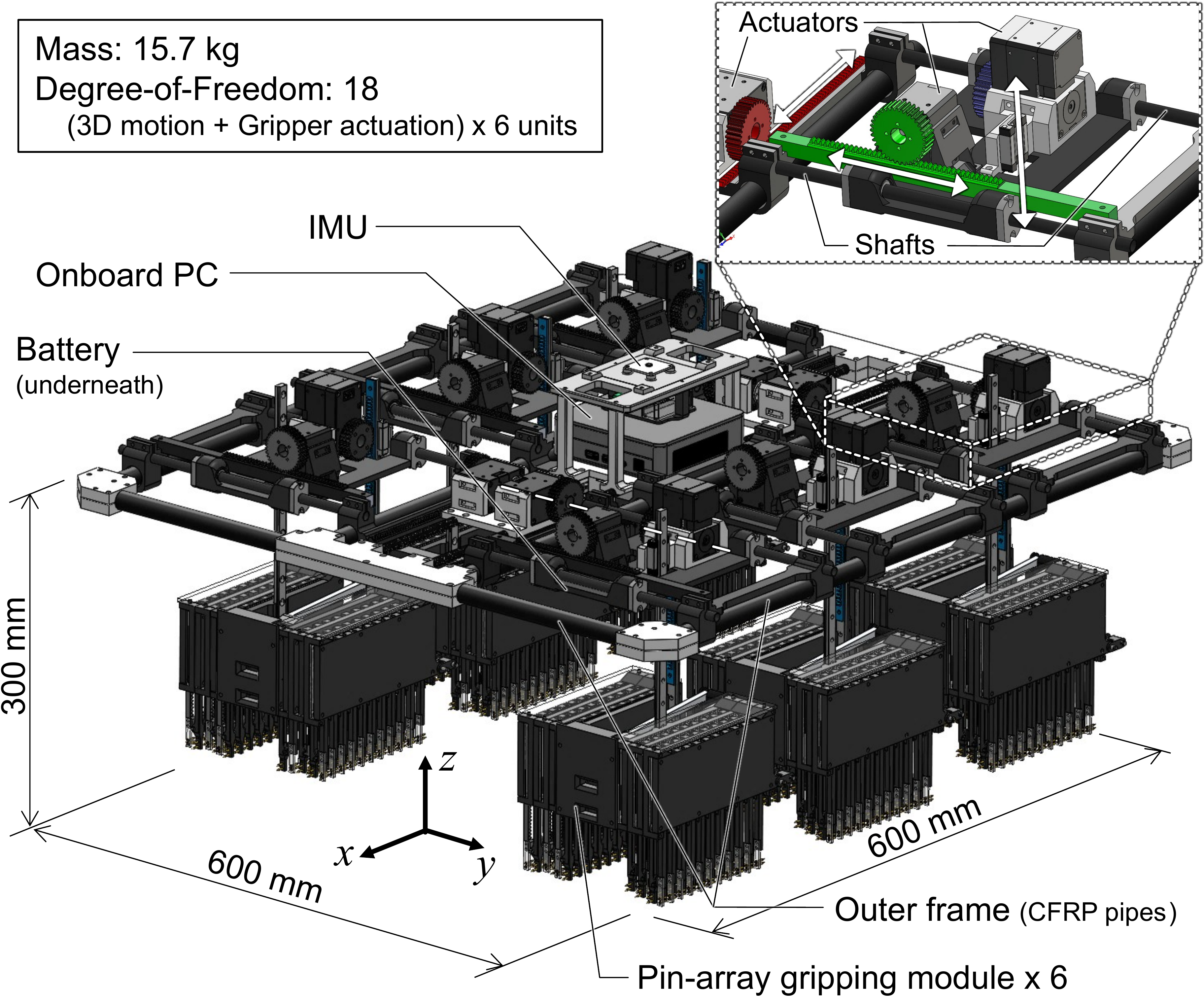}
  \caption{Hardware details of the developed pin-array structured mobile robot, consisting of six gripping units.}
  \label{fig:pin-array_robot}
\end{figure}
\begin{figure}[t]
  \centering
  \includegraphics[width=\linewidth]{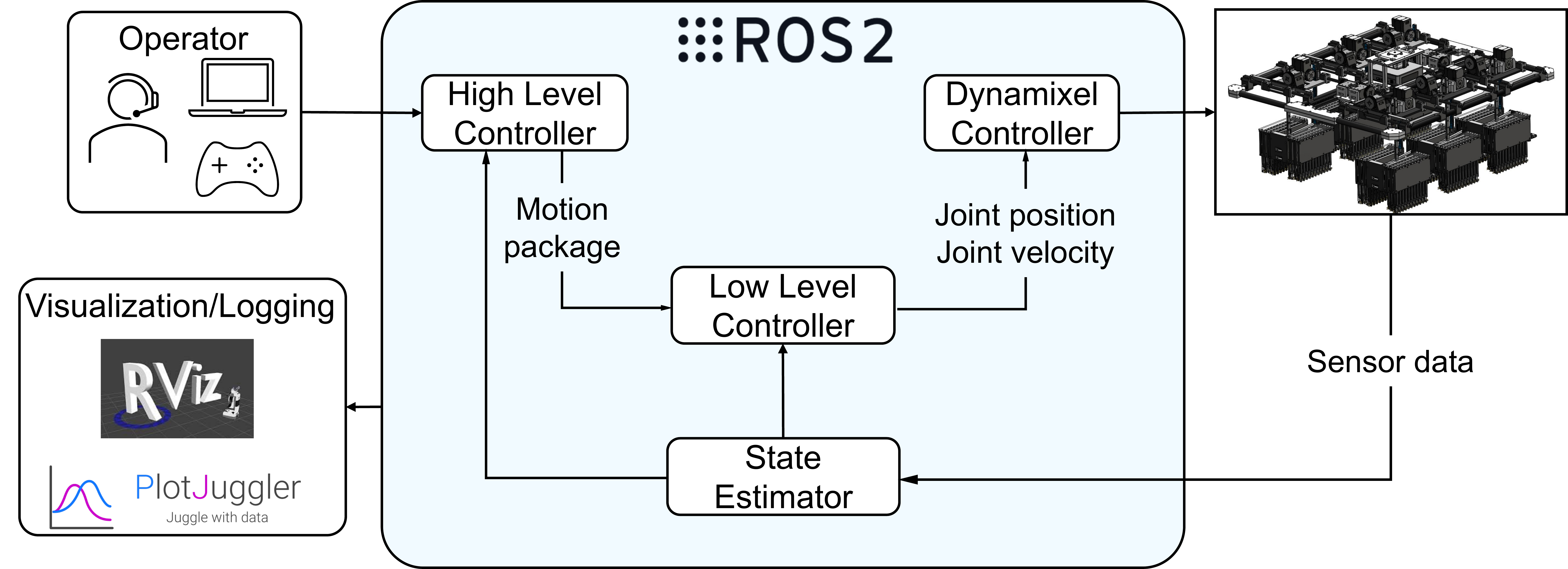}
  \caption{Software system diagram of the robot.}
  \label{fig:software_system}
\end{figure}
In this study, By exploiting the pin-array gripper, a novel mobile climbing robot is realized.

\subsection{Robot Hardware}
The robot developed is equipped with six gripping units, each adopting a pin-array gripper configuration. Each unit is designed with a three-degree-of-freedom prismatic joint structure. The mass of each unit is 1.4\;kg, resulting in a total robot mass of 15.7\;kg. An overall view of the robot is presented in \fig{fig:pin-array_robot}.

As illustrated in the top part of \fig{fig:pin-array_robot}, the upper section of the gripper unit houses a dedicated driving module. 
This module employs two actuators that enable translational motion of the gripper along either the $y$ or $z$ axis, 
while additional actuators mounted on the robot frame provide full positioning capability within the $x,y,z$ workspace. 
Each axis utilizes a rack-and-pinion linear drive mechanism, ensuring vertical pressing against the ground—a critical 
constraint arising from the characteristics of the pin-array gripper.

The list of hardware components integrated into the robot is provided in Table~\ref{tab:elec}. 
An Intel NUC 13 Pro serves as the onboard computer, capable of executing all control modules locally. 
Remote monitoring and operation are achieved by establishing an SSH connection between the onboard NUC 
and the operator's workstation. Power is supplied via a hybrid configuration: the actuators are driven by an 
external power source, whereas the onboard computer and sensors draw power from the integrated lithium-ion battery.
\begin{table}[hb]
    \scriptsize
    \centering
    \caption{Employed hardware components of the robot.}
    \label{tab:elec}
    \begin{tabular}{llr}
        \hline 
        Component & Model & Quantity\\
        \hline 
        Robot & & \\ 
        \hline
        Embedded Computer & Intel NUC 13 Pro (NUC13ANKi7) & 1\\
        IMU & RT-USB-9axisIMU2 & 1\\
        USB-Serial Converter & USD2 & 1\\
        Lithium-ion Battery & Anker Prime Power Bank 27650mAh & 1\\
        \hline 
        Pin-Array Gripper Unit & & \\ 
        \hline
        Joint Actuator & Dynamixel XM540-W270R & 6\\
        Joint Actuator & Dynamixel XM430-W350R & 12\\
        Gripper Actuator & Dynamixel XM430-W350R & 6\\
        \hline 
    \end{tabular}
\end{table}
\subsection{Software Integration}
\begin{figure*}[t]
  \centering
  \includegraphics[width=\linewidth]{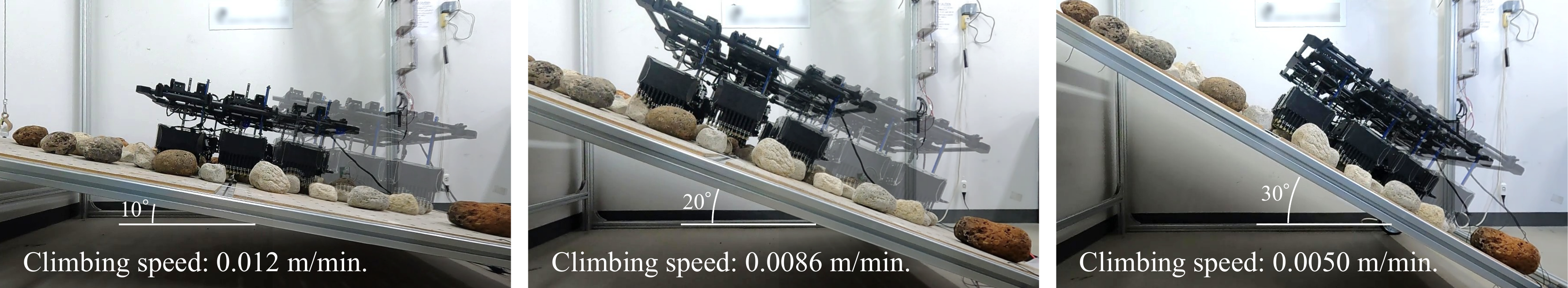}
  \caption{Indoor climbing experiments conducted on an artificial rock field. The robot was tested on slopes of 10$^{\circ}$, 20$^{\circ}$, and 30$^{\circ}$ inclination (from left to right) to evaluate its climbing performance.}
  \label{fig:indoor_locomotion_test}
\end{figure*}
The robot's software system was developed in ROS~2~\cite{ros2} and C++ using a modular, event-driven architecture (\fig{fig:software_system}). Operator commands are processed by a High-Level Controller (HLC), converted to unit-specific instructions by a Low-Level Controller (LLC), and transmitted to the actuators via the Dynamixel Controller, which also provides encoder and current feedback.

The system supports two modes: (a) manual, allowing direct control of individual units for movement and grasp/release actions, and (b) semi-autonomous, where directional commands (e.g., forward, left) trigger continuous execution of predefined locomotion patterns until updated.

The architecture consists of three primary modules---\textit{Control}, \textit{State Estimation}, and \textit{Motion Planning}---each implemented as an independent ROS~2 node communicating asynchronously via topics, services, and actions. The \textit{State Estimator} fuses encoder and IMU data to estimate the robot pose and grasp states. 

This modular design improves flexibility, supports parallel execution for real-time responsiveness, and facilitates future extensions such as terrain perception and adaptive grasp-force control.

\section{Experiments}
This section presents climbing experiments conducted to evaluate the locomotion performance of the robot equipped with the developed pin-array gripper. Experiments were carried out in two environments: (A) an indoor field with configurable inclination angles and grasping points, and (B) an outdoor natural landscape containing irregular rock surfaces. In both cases, the actuators were powered externally.

The robot employed a backward wave gait, which provides high static stability in hexapod robots and allows each leg to maintain its maximum stride. Although the robot was teleoperated, no adjustments were made to the driving sequence or grasping positions based on environmental conditions; all legs executed identical repetitive motions during climbing.


\subsection{Static Experiments}
To evaluate the fundamental performance of the developed robot, we experimentally determined the limits of its static stability. In this evaluation, we verified the maximum inclination angle at which the robot could maintain a stable posture in a static state with all grippers engaged. As shown in Fig. 1 (bottom left), the robot successfully maintained its posture up to an inclination of $65^\circ$. At this angle, mechanical failure occurred in the form of claw tips disengaging from the polycarbonate surface at multiple points; therefore, we defined this as the critical limit angle.

Let $M = 15.7$\,kg be the total mass of the prototype, $g = 9.81$\,m/s$^2$ be the gravitational acceleration, $F = 21.1$\,N be the holding force obtained from single-gripper characterization tests, and $n = 6$ be the total number of grippers. 
Under static equilibrium conditions, the relationship between the climbing angle $\theta$ and these parameters is given by:
\begin{equation}
    Mg \sin \theta = nF
\end{equation}
Thus, the theoretical maximum climbing angle $\theta_{\text{theory}}$ is calculated as follows:
\begin{equation}
    \theta_{\text{theory}} = \arcsin(0.822) \approx 55.3^\circ
\end{equation}
The experimental maximum angle of 65$^\circ$ exceeded the theoretical prediction of 55.3$^\circ$. This discrepancy can be attributed to two main factors. First, the theoretical model in Eq.~(1) only considers the active holding force $F$ measured in single-gripper tests, neglecting the passive frictional forces between the robot’s structure and the climbing surface. Second, as the inclination angle increases, the component of the robot's weight perpendicular to the slope enhances the mechanical interlocking of the claw tips into the polycarbonate surface, effectively increasing the actual holding force per unit beyond the value obtained in isolated characterization. These results demonstrate that the prototype possesses a higher safety margin and greater robust climbing stability than a simplified static model suggests.

\subsection{Indoor Experiments}
To demonstrate the climbing capability of the developed robot, multiple experiments were conducted on an artificial rock field with varying slopes~\cite{uno2024testfield}. Figure~\ref{fig:indoor_locomotion_test} shows the robot ascending the testbed. On a $10^{\circ}$ slope (left), the robot successfully traversed approximately 350\;mm in 30\;min., corresponding to a speed of 0.012\;m/min. On a $20^{\circ}$ slope (center), it advanced about 500\;mm in 58 minutes, confirming the effectiveness of the gripping mechanism. Furthermore, on a $30^{\circ}$ slope (right)---a condition where wheeled or tracked locomotion generally fails---the robot succeeded in climbing approximately 180\;mm within 36\;min. These results highlight the utility of the system as a climbing robot. In particular, the experiments demonstrated that, by combining a compliant gripper mechanism with the robot, effective climbing on unknown terrains can be achieved using only simple repetitive control sequences.

As the inclination angle increased, the climbing speed decreased. This was mainly because the slope component of the robot’s weight caused the claws to embed more deeply into the rock surface. To ensure stable movement, the $z$-axis velocity of each unit was reduced, which in turn limited the overall locomotion speed. These observations guide two further improvements of the robot: 1) hardware modifications to increase the stride length of each unit, and 2) software strategies in which the lifting motion is preceded by a slight movement along the slope direction, thereby loosening the claw embedding to facilitate disengagement and prevent unintended snagging. Implementing such improvements could enable faster and more reliable climbing performance under steeper conditions.

\subsection{Outdoor Experiments}
\begin{figure*}[t]
  \centering \vspace{5mm}
  \includegraphics[width=\linewidth]{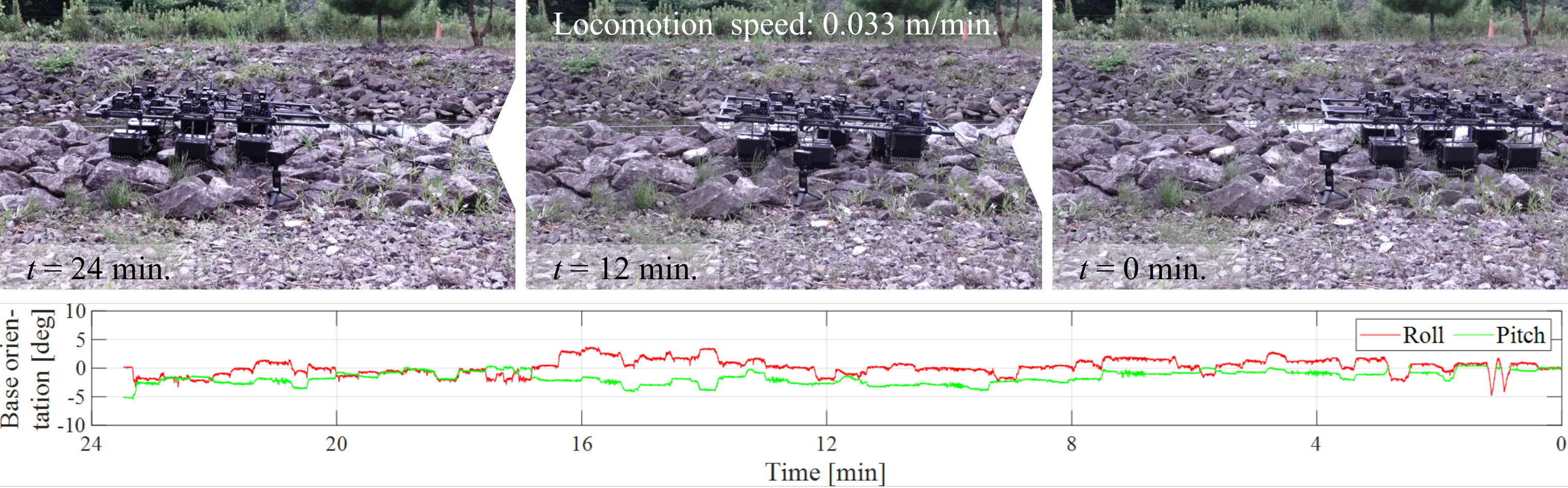}
  \caption{Locomotion experiments in an outdoor natural rocky environment. Snapshots showing the robot's traveling history from right to left (top) and the measured robot's attitude fluctuations throughout the locomotion (bottom). The robot demonstrates stable traversal of the natural rocky terrain while maintaining a consistent posture (within $\pm5^\circ$ in roll and pitch), owing to its highly terrain-adaptive locomotion.}
  \label{fig:outdoor_locomotion_test}
\end{figure*}
Unlike the controlled indoor environment, the target operational environment of the robot is inherently more complex and less predictable. To evaluate the robot's environmental adaptability and robustness, experiments were conducted in a natural rocky environment (see the bottom-right picture in \fig{fig1}), consisting of natural rocks approximately 10--20\;cm in height that create irregular terrain with heterogeneous friction characteristics due to the presence of sand and plants.

The robot was manually operated to traverse a straight path of roughly 1\;m. To increase the robot's locomotion speed in this experiment, the guide rails of the $x$-axis drive module were extended, allowing the stride length to increase from the original 90\;mm to 200\;mm.

\fig{fig:outdoor_locomotion_test} shows the outdoor experiment. The robot successfully traversed approximately 0.8\;m over 24\;minutes while adapting flexibly to the irregular terrain. This result demonstrates the high capability of the proposed robot system to adapt to unstructured environments, which could not be fully evaluated through indoor experiments alone. In particular, the six-unit configuration provided superior redundancy, enabling the system to absorb local instabilities such as slight displacements or unexpected inclinations of individual rocks—conditions that differed significantly from those in the controlled indoor field. This redundancy played a decisive role in maintaining overall stability throughout locomotion.

These findings confirm that the developed six-unit climbing robot can exhibit robust locomotion performance not only in controlled settings but also in complex and uncertain outdoor environments. Crucially, the experiments validate the robot's perception-free locomotion capability, where stable climbing is achieved through the intrinsic passive adaptation of the grippers rather than active terrain sensing.
\section{Conclusion}
This paper presented a novel mobile robot with pin-array grippers for stable climbing on steep, natural rocky terrain. The key contribution is achieving high stability through mechanical intelligence, where the gripper's physical compliance eliminates the need for complex, computationally expensive sensing and planning for grasping point detection. This allows the robot to grasp irregular surfaces through simple, uniform motions. Experimental results demonstrated stable locomotion in both controlled indoor settings and uncertain outdoor environments, confirming the robustness of our perception-free climbing approach.

Future work will first focus on overcoming identified speed limitations through a decoupled software strategy. While maintaining deliberate $z$-axis motion to manage spring reaction forces, we will accelerate movements along the progression axis, integrated with a pre-lifting displacement strategy for smoother claw disengagement. Furthermore, we aim to optimize the gripper for weight reduction and improved configuration. A lightweight design would facilitate integration into existing climbing robots as an end-effector, contributing to the realization of perception-free systems that do not require complex environmental sensing. This versatility also suggests potential for applications in agriculture or logistics. Ultimately, integrating autonomous control will enable real-time adaptation to diverse environments, achieving both high-speed, stable climbing and intelligent manipulation in unstructured terrains.

\section*{Acknowlegment}
The authors would like to express their sincere gratitude to Takuya Kato, Kenta Iizuka, and Ryuya Matsuoka for the effective discussion and the invaluable support in the experiments.

\bibliography{./IEEEabrv,bibliography.bib}

\end{document}